\documentclass[runningheads]{llncs}

\usepackage{eccv}

\usepackage{eccvabbrv}
\usepackage{microtype}
\usepackage{graphicx}
\usepackage{booktabs}
\usepackage{amsmath}
\usepackage{amssymb}
\usepackage{mathtools}
\usepackage{algorithm}
\usepackage{algorithmic}
\usepackage[accsupp]{axessibility}
\usepackage{hyperref}
\usepackage{orcidlink}

\spnewtheorem{assumption}{Assumption}{\bfseries}{\itshape}

\begin{document}

\title{Diffusion Path Alignment for Long-Range Motion Generation and Domain Transitions}

\author{
Haichao Wang\inst{1}\thanks{These authors contributed equally to this work.}\and
Alexander Okupnik\inst{2}\textsuperscript{$\star$}\and
Yuxing Han\inst{1}\and
Gene Wen\inst{3}\and
Johannes Schneider\inst{2}\and
Kyriakos Flouris\inst{4}
}

\authorrunning{H.~Wang et al.}

\institute{
Shenzhen International Graduate School, Tsinghua University, China
\email{\{Wanghc23,yuxinghan\}@sz.tsinghua.edu.cn}
\and
Institute of Information Systems, Universität Liechtenstein, Liechtenstein
\email{\{alexander.okupnik,johannes.schneider\}@uni.li}
\and
Department of Computer Science, New York University, USA
\email{jw9263@nyu.edu}
\and
MRC Biostatistics Unit, University of Cambridge, United Kingdom
\email{kyriakos.flouris@mrc-bsu.cam.ac.uk}
}

\maketitle

\begin{center}
    \centering
    \includegraphics[width=0.9\textwidth]{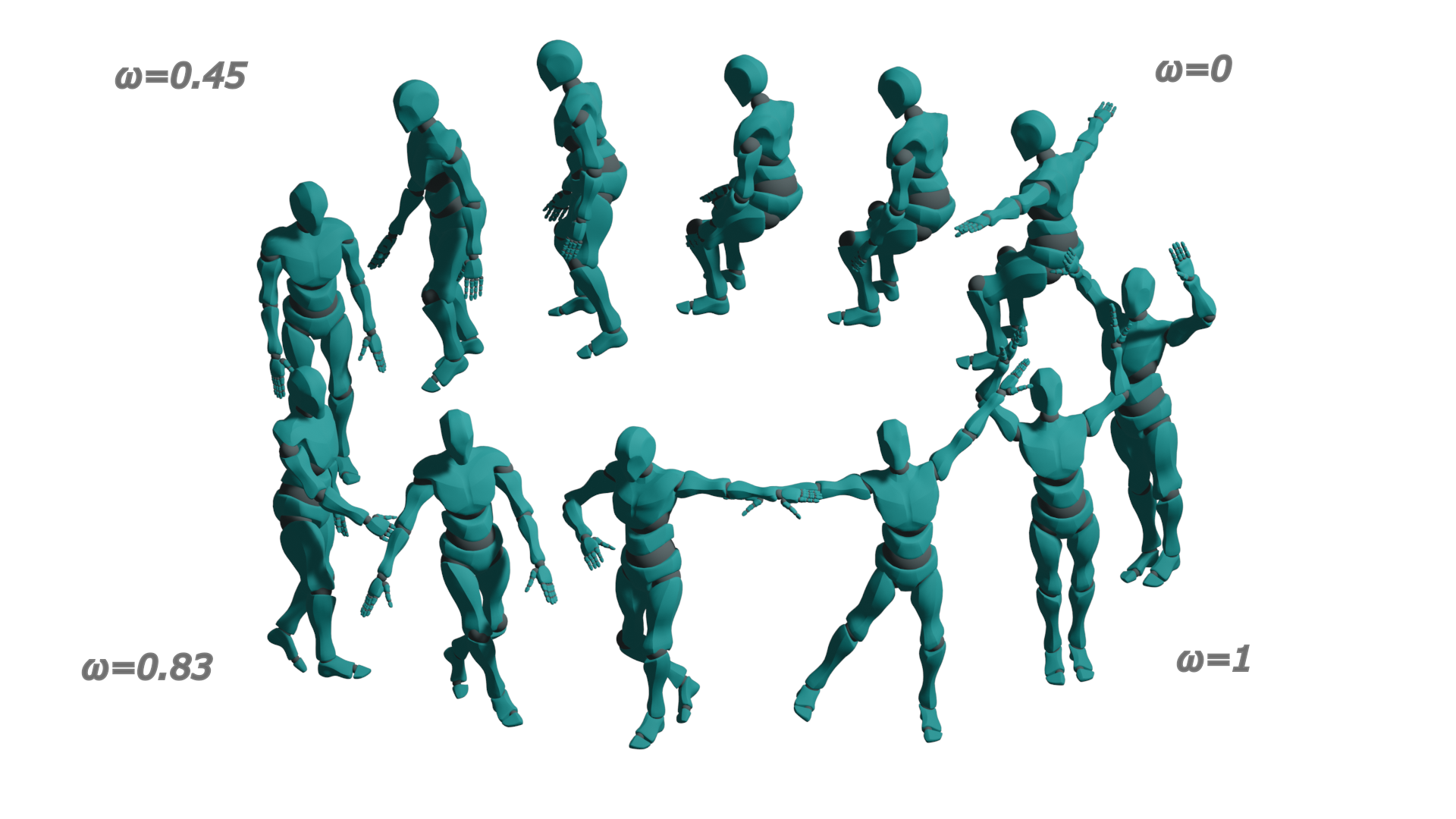} 
    \captionof{figure}{We propose Movement Diffusion Path Alignment (M-DPA) for controlled long-range motion generation. Our method optimizes segment-wise mixing coefficients between paired diffusion conditions $\omega$ at inference time, minimizing a control energy objective derived from stochastic optimal control. Hard stitching constraints enforce exact temporal continuity between consecutive motion segments. Example illustrates the transition from general movement $\omega=0$ domain to dance $\omega=1$.}
    \label{fig:two_domain_transport}
\end{center}

\begin{abstract}
Long-range human movement generation remains a central challenge in computer vision and graphics. Generating coherent transitions across semantically distinct motion domains remains largely unexplored. This capability is particularly important for applications such as dance choreography, where movements must fluidly transition across diverse stylistic and semantic motifs. We propose a simple and effective inference-time optimization framework inspired by diffusion-based stochastic optimal control. Specifically, a control-energy objective that explicitly regularizes the transition trajectories of a pretrained diffusion model. We show that optimizing this objective at inference time yields transitions with fidelity and temporal coherence. This is the first work to provide a general framework for controlled long-range human motion generation with explicit transition modeling.
\end{abstract}

\section{Introduction}
Despite recent advances in generative modeling for human motion, producing long-range, temporally coherent movement remains a major challenge. In particular, generating transitions between semantically distinct motion domains has received limited attention. Addressing this problem is crucial for applications such as dance choreography, where creativity is essential and sequences must fluidly evolve across diverse stylistic and semantic patterns while simultaneously maintaining continuity and realism.

 Practical systems often require controllable generation, not only realism. In interactive scenarios, motion must remain temporally coherent with previously generated movement while adapting to changing semantic objectives. For example, an interaction scenario may require transitioning between different movement domains when the current motion falls outside the desired behavioral regime. This raises a fundamental question: how can transitions between distinct motion domains be modeled in a principled manner when the underlying generative model is trained only on separate movement classes?

Diffusion models have recently provided powerful tools for temporally coherent motion synthesis. Diffusion-based approaches achieve state-of-the-art results in unconditional and conditional generation, enabling applications such as text-driven motion generation, editing, and motion completion. Architectures such as Motion Diffusion Model (MDM) \cite{tevet2022mdm} and EDGE \cite{tseng2023edge} demonstrate that diffusion processes combined with classifier-free guidance (CFG) can produce expressive motion while enabling flexible multi-modal control.

Despite these successes, most practical control mechanisms in diffusion-based motion generation rely on \emph{heuristically chosen parameters}. In particular guidance strengths are selected manually. When guidance is applied aggressively---for example to enforce sampling from a specific class---samples may drift away from the learned data manifold, leading to reduced fidelity and diversity as discussed in \cite{jin2026stagewisedynamicsclassifierfreeguidance, park2025temporalalignmentguidanceonmanifold}. In motion generation this often manifests as jitter, instability, or implausible motion trajectories. While heuristic schedules for guidance weights are widely used in practice, there is currently little principled understanding of how such parameters should be selected in the presence of long-range temporal constraints.

Recent work has begun to analyze diffusion models through the lens of \emph{stochastic optimal control} (SOC). Diffusion sampling can be formulated as solving a stochastic differential equation whose drift corresponds to a learned score field. From this perspective, conditional sampling can be interpreted as steering a reference stochastic process toward a target distribution while minimizing a quadratic control cost. Importantly, these analyses reveal that deviations from unconditional diffusion dynamics incur a control-energy penalty, and excessive steering can degrade sample quality by pushing trajectories off the data manifold.

This control-theoretic view suggests a natural reinterpretation of commonly used guidance mechanisms. Classifier-free guidance, blending weights for temporal coherence, or other test-time control coefficients can be seen as \emph{restricted control inputs} applied along denoising time. However, in existing methods these inputs are typically fixed or heuristically scheduled. We argue that these coefficients should instead be treated as \emph{optimization variables}, chosen to minimize a well-defined control-energy objective while satisfying task-specific constraints.

In this work, we adopt a control-theoretic perspective on diffusion models to address an underexplored problem in motion generation: coherent transitions between distinct movement domains. First, we introduce a parameterized transition mechanism within a pretrained diffusion model using classifier-free guidance. Second, we derive a control-energy objective that optimizes these parameters while enforcing boundary conditions such as temporal continuity and source–target domain alignment. Third, we demonstrate that the proposed approach outperforms commonly used heuristic schedules in terms of motion fidelity and diversity, highlighting the benefits of principled inference-time optimization for problems that are typically addressed through manual tuning.

\section{Related Work}

\paragraph{Motion continuation.}
Early work on human motion generation focused on motion continuation using autoregressive recurrent architectures such as RNNs, LSTMs, and GRUs~\cite{fragkiadaki2015recurrent,jain2016structural,martinez2017human}. These models predict future poses conditioned on a short motion prefix, establishing the canonical formulation of motion synthesis as sequential forecasting. While effective for short-term prediction, recurrent approaches suffer from drift and error accumulation over long horizons due to compounding autoregressive errors and limited ability to model multi-modal futures~\cite{martinez2017human}. Transformer-based architectures later improved long-range modeling by capturing global temporal dependencies in motion sequences~\cite{aksan2021spatiotemporal,mao2020learning}.

\paragraph{Diffusion models for motion generation.}
Diffusion models have recently emerged as a powerful paradigm for motion synthesis due to their stability and ability to model complex multi-modal distributions. The Motion Diffusion Model (MDM) adapts denoising diffusion probabilistic models to human motion using a Transformer backbone, enabling unconditional generation as well as text- or condition-driven motion synthesis~\cite{tevet2022mdm}. Diffusion models naturally support editing tasks such as motion completion and in-betweening by enforcing constraints during the reverse diffusion process. Subsequent works further demonstrate the flexibility of diffusion models for motion editing and constrained generation~\cite{tseng2023edge,zhang2023motiondiffuse}.

\paragraph{Motion diffusion as a generative prior.}
A key conceptual development is the use of pretrained diffusion models as reusable motion priors that can be composed at inference time. Shafir \etal~\cite{shafir2023prior} introduce several composition strategies based on a fixed MDM prior, including sequential composition for long motions (DoubleTake), parallel composition for multi-person interaction (ComMDM), and model composition for fine-grained control (DiffusionBlending). These approaches highlight the versatility of pretrained diffusion priors for composing motion generation tasks. In contrast, our work focuses on exploiting classifier-free guidance within a single diffusion model trained on multiple movement classes to address the problem of coherent domain transitions.

\paragraph{Long-range motion generation.}
EDGE~\cite{tseng2023edge} extends diffusion-based motion generation to long-range music-conditioned dance synthesis using a Transformer backbone and audio features extracted from Jukebox~\cite{dhariwal2020jukebox}. Conditioning is incorporated through feature-wise modulation using FiLM-style affine transformations~\cite{perez2018film}, enabling expressive choreography aligned with music over extended time horizons. From a stochastic optimal control perspective, such modulation mechanisms can be interpreted as injecting control signals into the generative dynamics, where modulation coefficients determine how strongly conditional information perturbs the underlying diffusion process. Due to its strong performance in long-range motion generation under guidance control, we adopt EDGE as the generative backbone in our experiments.

\paragraph{Stochastic optimal control and diffusion trajectory optimization.}
Recent theoretical work establishes formal connections between diffusion models and stochastic optimal control. Berner \etal~\cite{berner2024oc} interpret diffusion sampling as steering a reference stochastic process while minimizing a control cost defined through path-space KL divergence. Building on this perspective, Diffusion Trajectory Matching (DTM) formulates inference-time guidance as a variational control problem over diffusion trajectories, enabling principled steering of pretrained models without retraining~\cite{pandey2024dtm}. While the authors of DTM introduce a similar paradigm, the application area of motion domain transition and temporal coherence as a boundary condition demands a different approach.

HardFlow extends this idea to flow-matching models and addresses constrained trajectory generation through numerical optimal control and apply it to human-object interaction~\cite{li2025hardflow}. 

These works demonstrate the potential of control-theoretic formulations for guiding generative models. Our work builds on this perspective and applies trajectory-level control to the problem of motion domain transitions, optimizing guidance mixing coefficients to produce temporally coherent transitions between movement classes.

\section{Problem Formulation}

Let $\mathcal{X}_A$ and $\mathcal{X}_B$ denote two human motion domains, each defined as distributions over temporally ordered pose sequences. 
A motion sequence of length $T$ is represented as
\begin{equation}
\mathbf{x}_{1:T} = (\mathbf{x}_1, \dots, \mathbf{x}_T), \quad \mathbf{x}_t \in \mathbb{R}^d,
\end{equation}
where $\mathbf{x}_t$ encodes the human pose at time $t$.
Given a source motion prefix $\mathbf{x}^{A}_{1:T_s} \sim \mathcal{X}_A$ and a target motion domain $\mathcal{X}_B$, our goal is to generate a continuation $\hat{\mathbf{x}}_{T_s+1:T}$ such that:
\begin{enumerate}
    \item \textbf{Temporal coherence:} The full sequence $(\mathbf{x}^{A}_{1:T_s}, \hat{\mathbf{x}}_{T_s+1:T})$ is temporally smooth and physically plausible.
    \item \textbf{Domain transition:} For sufficiently large $t$, $\hat{\mathbf{x}}_t \sim \mathcal{X}_B$.
    \item \textbf{Progressive transformation:} The transition from domain $A$ to domain $B$ occurs gradually rather than abruptly.
\end{enumerate}
Formally, we seek a conditional generative model
\begin{equation}
p_\theta(\mathbf{x}_{T_s+1:T} \mid \mathbf{x}^{A}_{1:T_s}, \mathcal{X}_B),
\end{equation}
that jointly enforces sequence-level realism and cross-domain distribution alignment.

Existing approaches typically address either long-horizon motion generation within a single domain, or domain translation via distribution matching without explicit temporal consistency. 
In contrast, we address the unified problem of time-coherent domain transition in human motion sequences.

\section{Method}

\subsection{Guidance Between Paired Conditions}

We consider guidance between two conditioned diffusion models corresponding to conditions $c_0$ and $c_1$. Let
\(
\epsilon_\theta(x_t,t;c_0)
\)
and
\(
\epsilon_\theta(x_t,t;c_1)
\)
denote the predicted noise under each condition. We additionally denote by
\(
\epsilon_\theta(x_t,t;\varnothing)
\)
the unconditional noise prediction.

We define a mixed guidance prediction parameterized by a mixing coefficient $\omega(t)$:
\begin{equation}
\epsilon_\theta(x_t,t;\omega)
=
(1-\omega(t))\,\epsilon_\theta(x_t,t;c_0)
+
\omega(t)\,\epsilon_\theta(x_t,t;c_1).
\label{eq:paired_mixing}
\end{equation}

The control-induced deviation is always measured relative to the unconditional diffusion trajectory:
\begin{equation}
\Delta \epsilon_\theta(x_t,t;\omega)
=
\epsilon_\theta(x_t,t;\omega)
-
\epsilon_\theta(x_t,t;\varnothing).
\label{eq:delta_eps}
\end{equation}

\subsection{Terminal and Transient Objectives}
\begin{figure}[htbp]
    \centering
    \includegraphics[width=0.7\textwidth]{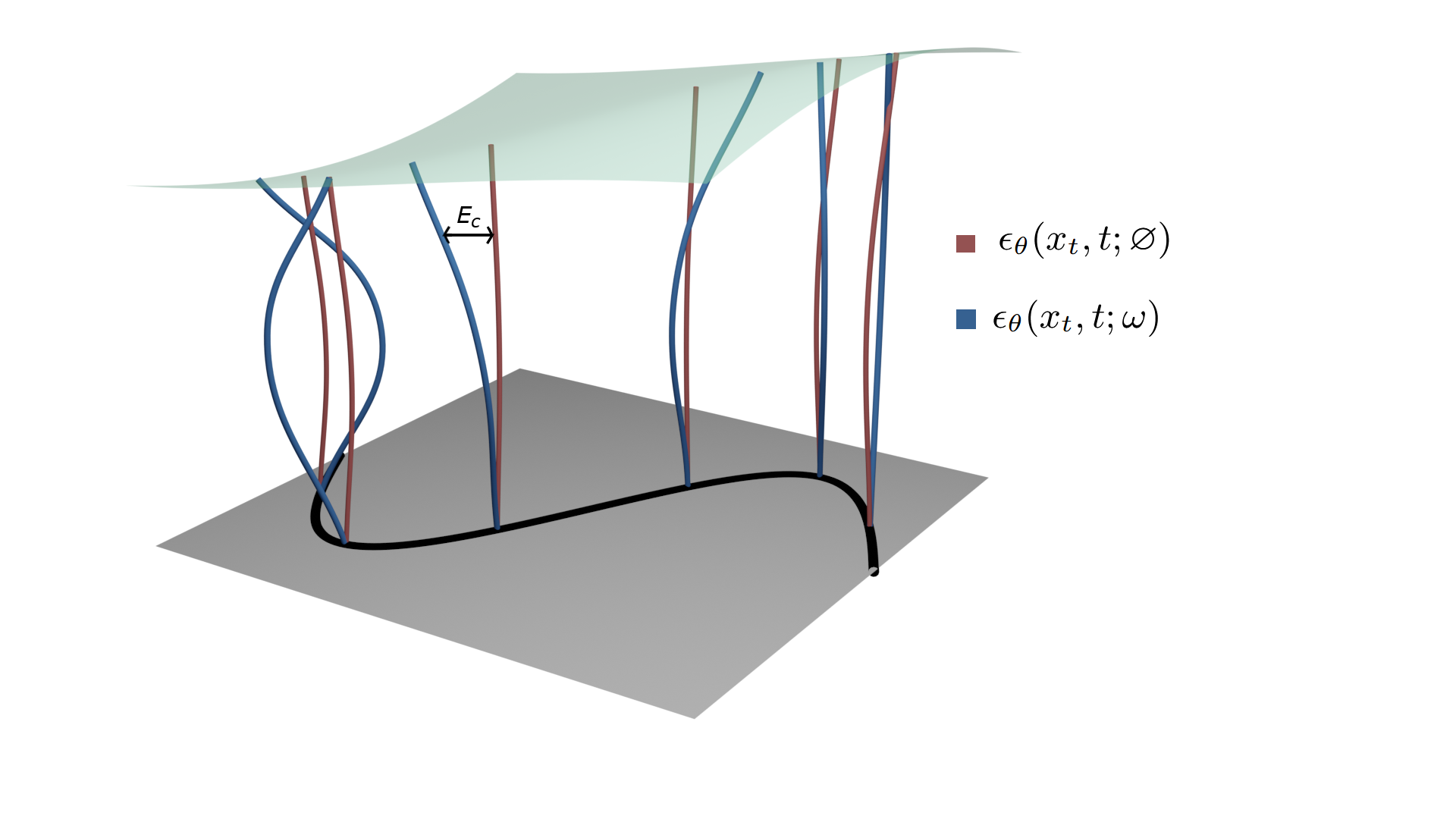} 
    \caption{Illustration of the guided denoising with additional control energy objective. Black line illustrations trajectory in latent space at $t=T$, red represents unconditional denoising trajectory with $\epsilon_\theta(x_t,t;\varnothing)$ and the blue line with mixed guided denoising $\epsilon_\theta(x_t,t;\omega)$. $E_c$ is the control energy to be minimized for the guidance mechanisms to be close to the unguided trajectory. }
    \label{fig:method_illustration}
\end{figure}
Following diffusion trajectory matching, guidance is derived from a stochastic optimal control objective consisting of a transient KL divergence and a terminal cost. Under $\epsilon$-prediction parameterization, the KL divergence between controlled and unconditional reverse transitions admits a quadratic form:
\begin{equation}
D_{\mathrm{KL}}
\big(
p(x_{t-1} \mid x_t, \omega)
\,\|\, 
p(x_{t-1} \mid x_t, \varnothing)
\big)
\;\propto\;
\|
\Delta \epsilon_\theta(x_t,t;\omega)
\|_2^2.
\label{eq:kl_eps}
\end{equation}
For a detailed derivation we refer the reader to the appendix \ref{app:transient}.

The terminal cost evaluated via Tweedie’s estimator $\hat{x}_0^{\,t}$ depends linearly on the predicted noise and therefore induces the same quadratic form. As a result, the terminal and transient objectives collapse into a single per-step control energy:
\begin{equation}
\mathcal{C}_t(\omega)
=
\lambda_t
\|
\Delta \epsilon_\theta(x_t,t;\omega)\|_2^2 +w_T\Phi(\hat{x}_{0,t})
,
\label{eq:collapsed_objective}
\end{equation}
with $\Phi$ representing the boundary constraints, $w_T$ a time-independent regulation parameter and where $\lambda_t$ absorbs noise-level– and sampler-dependent constants.

Thus, the mixing coefficient $\omega(t)$ fully parametrizes the deviation from the unconditional diffusion trajectory.

\subsection{Segmented Mixing Schedules}

We partition the diffusion trajectory into $K$ disjoint segments $\{\mathcal{T}\}_{k=1}^K$ with $\mathcal{T}$ being the denoising schedule and define a piecewise-constant mixing parameter
\begin{equation}
\omega_k(t),
\qquad
t \in \mathcal{T}.
\end{equation}

The total control energy decomposes additively across segments:
\begin{equation}
E_c
=
\sum_{k=1}^{K}
\sum_{t \in \mathcal{T}}
\lambda_t
\|
\Delta \epsilon_\theta(x_t,t;\omega_k(t)) 
\|_2^2+w_T\Phi(\hat{x}_{0,t}),
\label{eq:segmented_energy}
\end{equation}
with $\Phi$ ensuring the boundary condition of temporally consistent segment transitions and the first and last sample being drawn from the source and the target class respectively. Here, for samples with fixed length $S$, $\Phi$ from is defined by the L2 distance between overlapping regions of adjacent segments:
\begin{equation}
     \Phi(\hat{x}_{0,t}) = \sum_{k=1}^{K-1} \left\| \hat{x}_{0,t}^{(k+1)}[1:S/2] - \hat{x}_{0,t}^{(k)}[S/2:S] \right\|_2^2 
\end{equation}
We also perform deterministic alignment on the root translation channel for each segment: shifting the starting root position of the (k+1)-th segment to align with the ending root position of the k-th segment, thereby eliminating global displacement bias.

\subsection{Hard Stitching Constraints for Sequential Sampling}

We generate samples of fixed length $S$ and stitch consecutive samples to form longer trajectories. Let $x^{(i)}_{0:S}$ denote the $i$-th generated sample. We impose a hard continuity constraint:
\begin{equation}
x^{(i+1)}_{0:S/2}
=
x^{(i)}_{S/2:S}.
\label{eq:hard_stitch}
\end{equation}

\subsection{Long-Range Sampling Algorithm}




\begin{algorithm}[t]
\caption{Long-Range DDIM Sampling with M-DPA}\label{alg:online}
\begin{algorithmic}[1]
\REQUIRE Segments $K$, length $S$, conditions $(c_0,c_1)$, DDIM schedule $\{t_n\}_{n=1}^N$, optimization steps $J$
\ENSURE Long-range sample $\mathbf{x}_0 \in \mathbb{R}^{K \times S \times C}$
\STATE Sample $\mathbf{x}_T \sim \mathcal{N}(\mathbf{0},\mathbf{I})$
\FOR{$n=1$ \textbf{to} $N$}
    \STATE $\hat{\mathbf{x}}_0^{c_0},\;\hat{\mathbf{x}}_0^{c_1} \gets f_\theta(\mathbf{x}_t,t_n;\,c_0),\;f_\theta(\mathbf{x}_t,t_n;\,c_1)$
    \STATE Init $\boldsymbol{\omega}\gets[0,\;\sigma(\mathbf{z}),\;1]$ with learnable $\mathbf{z}\in\mathbb{R}^{K-2}$
    \STATE \COMMENT{optimize mixing weights}
    \FOR{$j=1$ \textbf{to} $J$} 
        \STATE $\hat{\mathbf{x}}_0^{(k)} \gets (1-\omega_k)\,\hat{\mathbf{x}}_0^{c_0}+\omega_k\,\hat{\mathbf{x}}_0^{c_1},\;\forall\,k$
        \STATE $\mathcal{C}_t \gets \lambda_t\!\sum_k\!\lVert\Delta\boldsymbol{\epsilon}_\theta^{(k)}\rVert_2^2 + w_T\!\sum_k\!\lVert\hat{\mathbf{x}}_{0,S/2:S}^{(k)}-\hat{\mathbf{x}}_{0,0:S/2}^{(k+1)}\rVert_2^2$ 
        \STATE $\mathbf{z}\gets\text{Adam}(\mathbf{z},\;\nabla_{\mathbf{z}}\mathcal{C}_t)$
    \ENDFOR
    \STATE Reconstruct $\hat{\mathbf{x}}_0$ with optimized $\boldsymbol{\omega}^\star$
    \STATE $\mathbf{x}_{t_{n+1}}\gets\text{DDIM}(\mathbf{x}_t,\,\hat{\mathbf{x}}_0,\,t_n,\,t_{n+1})$
    \STATE $\mathbf{x}_{0:S/2}^{(k+1)}\gets\mathbf{x}_{S/2:S}^{(k)},\;\forall\,k$ \COMMENT{Hard stitching Projection}
\ENDFOR
\RETURN $\mathbf{x}_0$
\end{algorithmic}
\end{algorithm}

\ref{alg:online} depicts the algorithm used in our experiments to exemplify the proposed method. For $K$ segments, $x_T$ were sampled form the normal distribution. For each denoising step $n \in N$, $N$ being the number of denoising steps, the mixing weights are uniformly initialized and then optimized with Eq. \ref{eq:segmented_energy} as objective. After obtaining the optimal mixing weights $\omega$, the predicted noise is stitched together. The estimated clean sample $\hat{x}_0$ is predicted and fed into the model to complete the DDIM step.

\section{Experiments}

\subsection{Datasets}

We train and evaluate on two motion datasets unified into a common representation.
\textbf{AIST++}~\cite{aist++} contains 1,408 sequences across 10 dance genres
performed by 30 dancers, totalling approximately 5.2 hours at 60\,FPS.
\textbf{HumanML3D}~\cite{humanml3d} comprises 14,616 motion sequences drawn
from AMASS~\cite{amass} and HumanAct12~\cite{humanact12}, covering a broad
range of everyday activities and totalling approximately 28.6 hours.

Both datasets are downsampled to 30\,FPS and sliced into fixed-length windows
of 5 seconds (150 frames) with a stride of 0.5 seconds, following the EDGE
setting~\cite{tseng2023edge}. We apply an 80/20 split at the sequence level prior to
windowing to prevent data leakage across partitions. A global normalizer is
constructed from training-split statistics only.

Motions are represented using the SMPL kinematic model~\cite{smpl} with 22
joints and the 6D continuous rotation parameterization~\cite{rot6d}
($22{\times}6=132$ dimensions). Each frame-level feature $x_t\in\mathbb{R}^{139}$ concatenates local joint rotations, global root translation (3 dims), and binary foot contact labels (4 dims) derived from velocity thresholds. AIST++ sequences (Y-up) are rotated to match the Z-up convention of AMASS.

For conditional generation we define 11 classes: the 10 AIST++ dance genres and a single aggregated class covering all HumanML3D motions, which serves as the source domain $c_0$ in our experiments.

\subsection{Experimental Setup}

Our model is built upon the EDGE Transformer backbone: an 8-layer decoder with latent dimension 512, 8 attention heads, and feed-forward size 1024. Style conditions are projected by a 2-layer Transformer encoder and injected via cross-attention and FiLM modulation~\cite{perez2018film} of the time embedding.

We use a Gaussian diffusion process with a cosine noise schedule over $T{=}1000$ timesteps, predicting the clean signal $x_0$ directly. The model is trained with the Adan optimizer~\cite{adan} at a learning rate of $4{\times}10^{-4}$, weight decay 0.02, condition drop probability 0.25, batch size 512, for 2000 epochs. The training objective is inherited from EDGE:
\begin{equation}
  \mathcal{L} = \lambda_{\text{rec}}\mathcal{L}_{\text{MSE}}
              + \lambda_{\text{vel}}\mathcal{L}_{\text{vel}}
              + \lambda_{\text{FK}}\mathcal{L}_{\text{FK}}
              + \lambda_{\text{foot}}\mathcal{L}_{\text{foot}},
\end{equation}
with $\lambda_{\text{rec}}{=}0.636$, $\lambda_{\text{vel}}{=}2.964$,
$\lambda_{\text{FK}}{=}0.646$, $\lambda_{\text{foot}}{=}10.942$.

At inference, we use DDIM~\cite{ddim} with 50 steps. The segment-wise mixing coefficients $\{\omega_k\}$ are optimized online at each denoising step via Adam with learning rate 0.01 for 20 gradient steps, minimizing the control-energy objective in Eq.~\ref{eq:segmented_energy}.


\subsection{Evaluation}

We evaluate generated motions using two complementary FID variants computed
from SMPL joint positions, following prior work on dance motion generation~\cite{li2023bailandopp}. 
FID\textsubscript{k} measures distributional similarity in a kinetic feature space, considering joint velocities and accelerations, capturing motion dynamics fidelity. 
FID\textsubscript{m} operates in a geometric feature space with joint positions and pairwise distances, capturing pose geometry fidelity.

To assess diversity, we report the average $\ell_2$ distance between randomly sampled pairs in the standardized feature space, measuring the spread of the generated distribution. 
All features are standardized using ground-truth statistics prior to evaluation, consistent with established evaluation protocols for long-range dance generation~\cite{li2023bailandopp}.

For each generation method, we produce 1000 motion clips and compare against 1000 ground-truth clips sampled from the test set with class-balanced sampling. All motion representations are first converted to 3D joint positions via SMPL forward kinematics before feature extraction.
For long-sequence generation methods, each generated long sequence of $K$ segments is assembled with 50\% overlap using linear cross-fade blending, then sliced back into fixed-length clips with a sliding window (stride = $S/2$) for evaluation. This ensures a fair comparison with single-segment ground truth.

\subsection{Results}

Table~\ref{tab:fid} reports FID and diversity scores for all baselines and our proposed M-DPA method. Our method achieves the lowest $\text{FID}_k$ and $\text{FID}_m$ among all transition strategies, outperforming the strongest heuristic baseline (Sine) by 13.79\% and 7.42\% on $\text{FID}_k$ and $\text{FID}_m$ respectively. This demonstrates that optimizing the mixing coefficient online yields generated motion distributions substantially closer to the ground truth. All interpolation baselines exhibit considerably high FID scores, confirming that fixed heuristic schedules fail to keep the generation staying on data manifold during cross-domain transitions.

Regarding diversity, we observe a mild increase in $\text{Div}_\text{k}$ but a decrease in $\text{Div}_\text{m}$. This discrepancy is expected as the two metrics probe different aspects of the motion manifold. First, the hard stitching/overlap constraints primarily regularize pose-level continuity, which reduces geometric variability and thus lowers $\text{Div}_\text{m}$. Second, the per-sample online optimization of $\omega$ alleviates the over-smoothing induced by a fixed interpolation schedule, yielding more realistic velocity/acceleration statistics and slightly higher $\text{Div}_\text{k}$ compared to interpolation baselines. Overall, our method trades a small amount of geometric diversity for improved temporal dynamics, while maintaining competitive realism.

\begin{table}
    \centering
    \caption{Comparison of baselines and our method. Bold indicates the best result.}
    \begin{tabular}{ccccc}
    \toprule
         & $\text{FID}_\text{k} \downarrow$ & $\text{FID}_\text{m} \downarrow$ & $\text{Div}_\text{k} \uparrow$ & $\text{Div}_\text{m} \uparrow$\\
        \midrule
        Ground Truth & 8.25 & 11.47 & 14.01 & 21.97 \\ 
        \midrule
        Linear   & 174.32 & 181.85 & 6.13 & 20.48 \\
        Sigmoid  & 171.85 & 178.83 & 6.08 & \textbf{20.91} \\
        Sine     & 148.48 & 148.47 & 7.04 & 20.80 \\
        M-DPA (Ours)   & \textbf{128.01} & \textbf{137.45} & \textbf{7.06} & 18.88 \\
    \bottomrule
    \end{tabular}
    \label{tab:fid}
\end{table}

\section{Discussion}

\subsection{Analysis of Motion Dynamics}
\label{sec:dynamics}
The mean and variance of acceleration and jerk from M-DPA are closer to ground truth than those of sine interpolation, indicating that our method restores the inherent dynamic diversity of natural motion in ground truth. The low variance in sine interpolation results from excessive smoothing that suppresses the vitality of motion.

\begin{table}[htbp]
\centering
\caption{Motion dynamics statistics across different settings.}
\label{tab:dynamics}
\begin{tabular}{lcccc}
\toprule
 & Accel Mean & Accel Var & Jerk Mean & Jerk Var \\
\midrule
Ground Truth & $4.526$ & $101.687$ & $138.975$ & $1.707\mathrm{e}{5}$ \\
\midrule
Sine         & $3.733$ & $73.151$  & $111.084$ & $1.076\mathrm{e}{5}$ \\
M-DPA       & $3.958$ & $91.427$  & $117.488$ & $1.449\mathrm{e}{5}$ \\
\bottomrule
\end{tabular}
\end{table}
\subsection{Qualitative Analysis of Motion}
\begin{figure}[htbp]
    \centering
    \includegraphics[width=\linewidth]{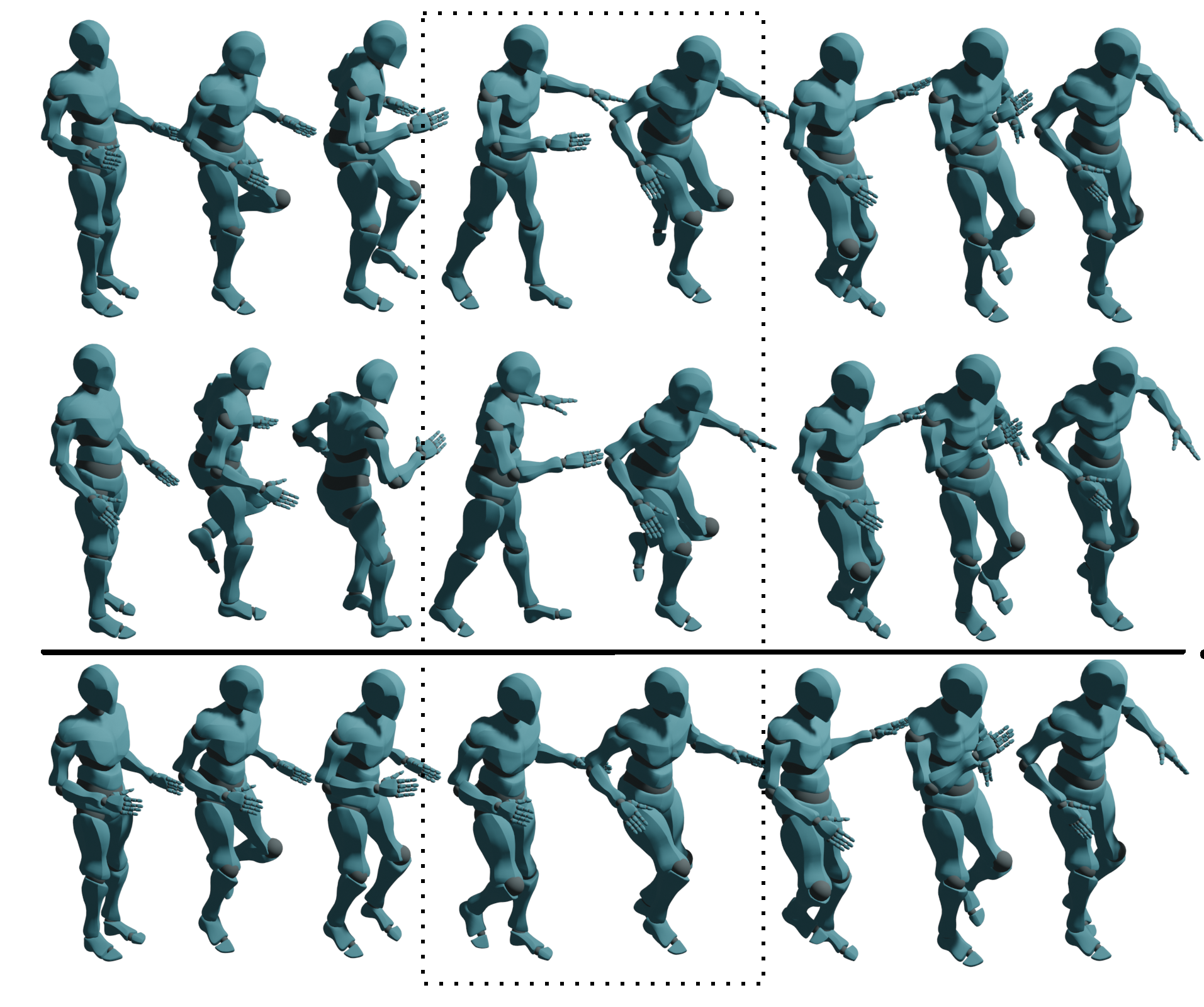}
    \caption{Subplots show generations of linear (first row), Sine (second row) and M-DPA (last row) generated movement samples. The dash lined bounding box marks the transition phase. While the heuristic methods like linear and sine interpolation introduces additional folding of the charakter, the M-DPA shows more coherent transitions.}
    \label{fig:comparison}
\end{figure}

Examining samples from the baseline of linear interpolating $\omega$ and our proposed optimization method yields smoother transitions between the source and target domain. Examples of it are displayed in Figure~\ref{fig:comparison}.

\subsection{Analysis of $\omega$ Optimization}

\begin{figure}[htbp]
    \centering
    \includegraphics[width=\linewidth]{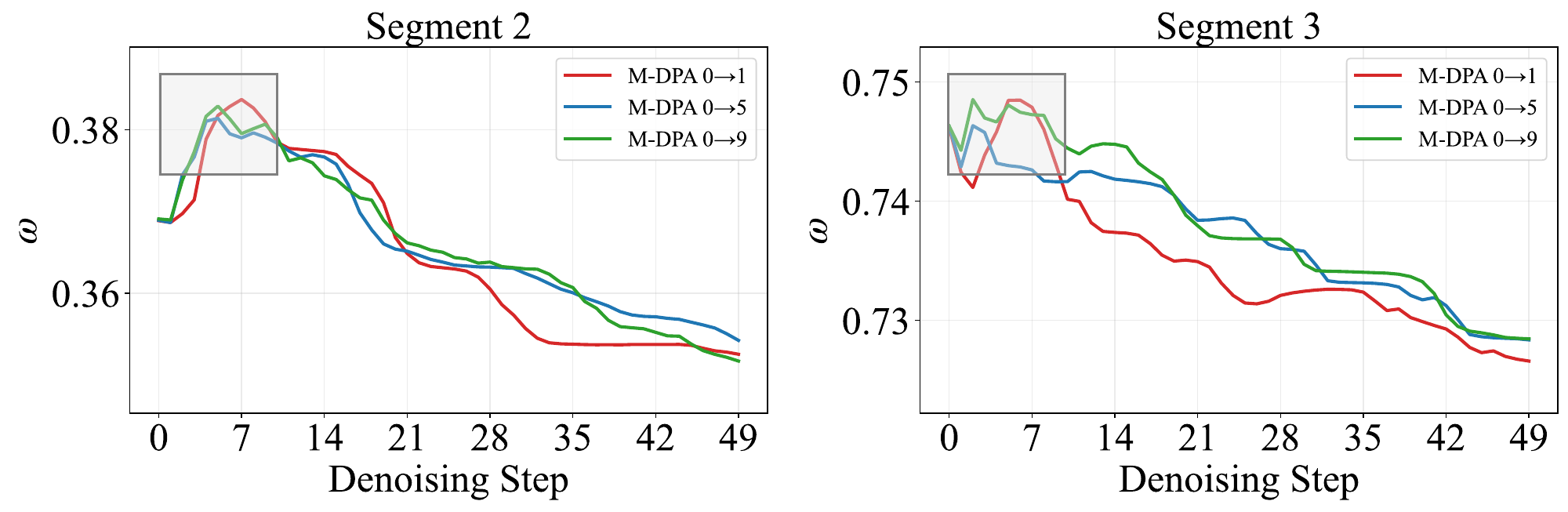}
    \caption{Subplots show the results of the $\omega$ optimization along different denoising steps for different class transitions. Both inbetween segments $2$ and $3$ show a similar pattern. $\omega$ peaks at denoising step $7$ and then monotonically decreasing, consistently for all the class transitions $0\rightarrow1$, $0\rightarrow5$, $0\rightarrow9$.}
    \label{fig:trajectory}
\end{figure}

To further understand the behavior of M-DPA, we analyze the learned mixing coefficients $\omega$ across denoising steps and motion segments. Figure~\ref{fig:trajectory} visualizes the optimized $\omega$ trajectories for three class transitions ($0\rightarrow1$, $0\rightarrow5$, $0\rightarrow9$) over segments 2 and 3.

A consistent pattern emerges across all transitions: $\omega$ rises sharply in the early denoising steps, peaks near step 7, and subsequently decreases monotonically. This behavior is because, during the high-noise phase in early denoising steps, the reverse process determines coarse semantic structure, necessitating strong guidance toward the target domain. As denoising progresses and fine-grained details are committed, the optimal control reduces guidance strength to avoid departing from the learned data manifold. The M-DPA objective automatically discovers this schedule, confirming that the control-energy formulation aligns with the intrinsic structure of the denoising process.

\subsection{Analysis on Control Energy}
\begin{figure}
    \centering
    \includegraphics[width=\linewidth]{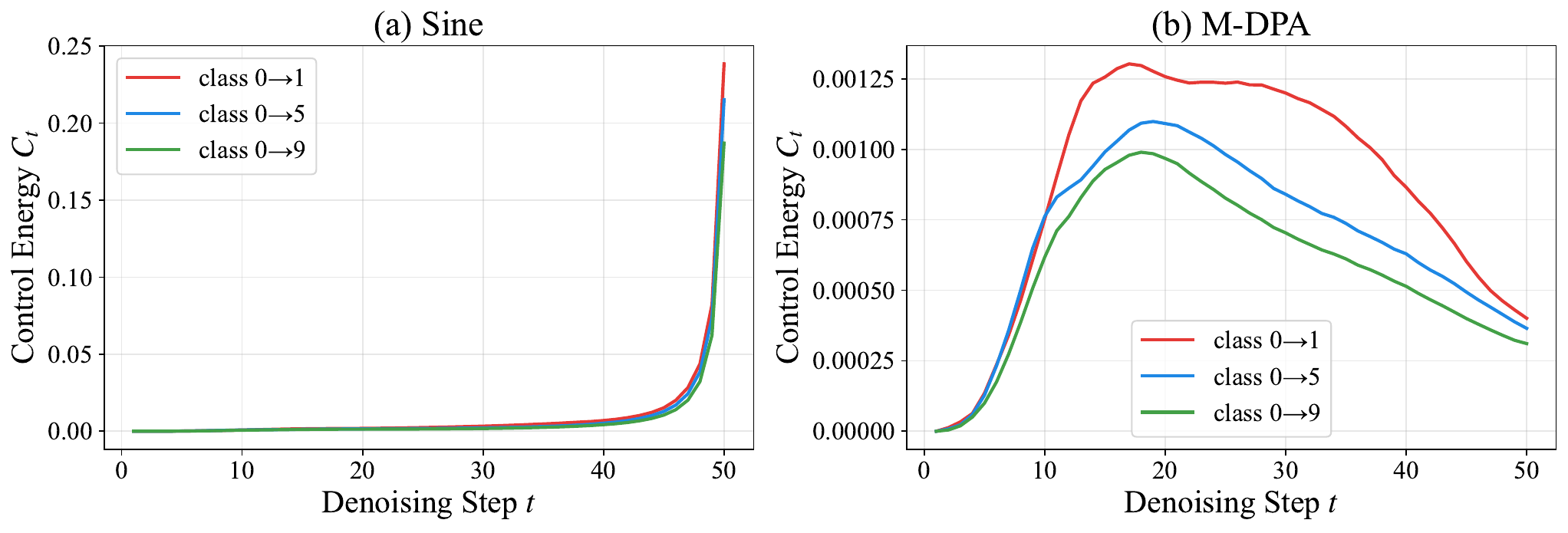}
    \caption{Evolution of control energy for both the Sine interpolation baseline and M-DPA.}
    \label{fig:energy}
\end{figure}

Control energy of Sine interpolation and M-DPA are recorded in Figure~\ref{fig:energy}. The control energy of the heuristic Sine strategy surges to nearly 0.25 during the late stages of denoising. In contrast, our method maintains control energy at extremely low levels throughout the entire denoising cycle, peaking only around 0.00125. This demonstrates that by online optimizing the mixing coefficient $\omega(t)$ during inference, our approach effectively minimizes the deviation between the guided trajectory and the unconditional diffusion trajectory.

For the Sine baseline, the control energy remains relatively stable during the initial to middle stages of denoising (steps 0–40), but exhibits an exponential surge in the latter stage (steps 40–50). The fine-grained features and local geometric structures of the generated samples gradually solidify as the denoising process progresses. At this stage, forcing the generation direction with fixed heuristic weights would compel the generated trajectory to deviate significantly from the learned data manifold, incurring extremely high control energy costs, manifesting as out-of-domain frames. Our M-DPA dynamically adjusts $\omega(t)$ to proactively reduce control intensity during the late denoising phase, thereby achieving semantic transitions while perfectly preserving the stability of generated samples on the data manifold.

\section{Conclusion and Outlook}

In this work, we showed how diffusion matching can be leveraged for controlled long-range motion generation, enabling coherent transitions between distinct semantic movement domains. By interpreting guidance as a trajectory-level control signal and optimizing it over time, we enable smooth and stable shifts from one movement style or action distribution to another.

Importantly, our approach operates purely at inference time and remains compatible with pretrained diffusion models. This makes it a practical tool for semantic motion editing, choreography transfer, and cross-domain animation. Our findings suggest that diffusion matching, when formulated at the trajectory level, provides a powerful mechanism for time-coherent semantic control in generative motion models.

\paragraph{Outlook.}
Several promising research directions emerge from this formulation. First, while we focused on classifier-free guidance mixing~\cite{ho2022classifierfree}, the proposed objective applies to broader guidance families, including RePaint~\cite{lugmayr2022repaint} and CFG++~\cite{chung2024cfgpp}. Evaluating the framework under these mechanisms may yield improved stability under stronger conditioning.

Second, replacing diffusion backbones with flow-matching~\cite{lipman2023flowmatching} or consistency models~\cite{song2023consistency} offers a natural extension. Recent work such as HardFlow~\cite{chen2024hardflow} suggests that trajectory-level optimal control can be combined with flow-based generative dynamics while preserving constraint satisfaction. Integrating diffusion trajectory matching (DTM)-style energy minimization into flow-matching samplers may further improve sampling efficiency. Related adaptive guidance strategies such as ratio-aware adaptive guidance (RAAG)~\cite{li2024raag} provide additional mechanisms for stabilizing strong conditioning.

Finally, our formulation suggests a more general view of inference-time hyperparameter optimization in generative modeling. Instead of manually tuning guidance schedules, future systems could automatically optimize control coefficients under task-dependent reward functions. This may enable principled composition of semantic commands, domain transitions, or multi-objective motion control without departing from the learned motion manifold.

\bibliographystyle{splncs04}
\bibliography{main}

\end{document}